\documentclass[sigconf]{acmart}
\usepackage{algorithm}
\usepackage{algpseudocode}
\usepackage{array}
\usepackage{stfloats}
\AtBeginDocument{\vbadness=10000}
\AtBeginDocument{%
  }

\copyrightyear{2026}
\acmYear{2026}
\setcopyright{rightsretained}
\acmConference[epiDAMIK @ KDD '26]{8th epiDAMIK ACM SIGKDD Workshop on Data-driven Decision Making for Public and Population Health}{August 10, 2026}{Jeju Island, Republic of Korea}
\acmBooktitle{Proceedings of the 8th epiDAMIK ACM SIGKDD Workshop on Data-driven Decision Making for Public and Population Health, August 10, 2026, Jeju Island, Republic of Korea}
\acmDOI{}
\acmISBN{}

\begin{document}

\title{Verifiable Knowledge Expansion through Retrieval-Grounded Formal Concept Analysis}

\author{Yujin Yang}
\email{yujinyang@hanyang.ac.kr}
\affiliation{%
  \institution{Hanyang University}
  \city{Seoul}
  \country{Korea}
}

\orcid{0009-0003-1379-2828}
\author{Heejung Lee}
\authornote{Corresponding author.}
\email{stdream@hanyang.ac.kr}
\affiliation{%
  \institution{Hanyang University}
  \city{Seoul}
  \country{Korea}
}

\renewcommand{\shortauthors}{Yang and Lee}

\begin{abstract}
  Ontology construction requires deciding which objects, attributes, and structural relations should be accepted as valid knowledge.
  Language models can propose such structures from text, but their outputs can still be unsupported or inconsistent.
  This paper proposes a retrieval-augmented small language model (SLM) framework that uses formal concept analysis (FCA) as a symbolic verification loop for knowledge expansion.
  Starting from seed attributes, FCA proposes implications over a growing formal context.
  A retrieval-grounded SLM oracle then validates each implication or returns a counterexample.
  The oracle also supports incidence judgments, consistency checks, and attribute proposals, making accepted implications, counterexamples, contradictions, and corrections inspectable.
  In a rare ataxia setting constructed from Orphadata resources, retrieval-grounded 10-seed runs obtain relation F1 of 0.29--0.52 and closure-based implication F1 of 0.22--0.30.
  Larger seed sets increase the number of evaluated implications and often improve implication F1.
  The lower implication scores reflect a stricter evaluation of derived implications, where one missed or extra relation can affect several implication judgments.
  Ablations show that incidence judgments in a fixed object--attribute setting can improve closure-based implication scores.
  However, identifying positive object--attribute pairs remains difficult even when the candidate objects and attributes are fixed.
\end{abstract}



\maketitle

\section{Introduction}
Ontologies turn domain knowledge into a structure that can be shared, queried, checked, and reused~\cite{gruber1993ontology,uschold1996ontologies,khadir2021ontology}.
In biomedical domains, for example, an ontology-like representation can connect rare disease objects to phenotype attributes and expose regularities among those phenotypes~\cite{robinson2008hpo,kohler2021hpo}.
However, building such structures manually is costly.
Domain experts must inspect source documents, encode object--attribute relations, and revise the structure when new evidence reveals missing or inconsistent knowledge~\cite{uschold1996ontologies,alaswadi2020automatic,khadir2021ontology}.
This makes ontology and knowledge graph construction a natural setting for language model assistance, but errors in domain specific knowledge make unverifiable generation risky~\cite{pan2024unifying,lo2024endtoend,huang2025hallucination}.
The key technical challenge is therefore to verify proposed knowledge before it becomes a structural commitment.
For example, the system may propose that a rare ataxia disease with \textit{Ataxia}, \textit{Cerebellar atrophy}, and \textit{Tremor} should also have \textit{Dysarthria}.
This rule is rejected if even one disease has the first three phenotypes but lacks \textit{Dysarthria}.
Ontology construction therefore needs a procedure that asks targeted structural questions and incorporates counterexamples when a proposed regularity fails~\cite{ganter2016conceptual}.
Formal concept analysis (FCA) provides such a procedure.
Its attribute exploration procedure proposes implications over a formal context and requires a counterexample when an implication is invalid~\cite{ganter1999fca,ganter2016conceptual}.

The proposed framework combines FCA with retrieval-grounded SLM judgments for rare ataxia diseases.
The task is to build a disease--phenotype formal context over standardized HPO labels such as \textit{Ataxia}, \textit{Tremor}, and \textit{Dysarthria}.
The practical question is whether a small seed attribute set can activate useful parts of the controlled phenotype attribute set while preserving verifiable incidence judgments and implication checks.
Across three small language models (SLMs), the system expands the context over 20 rounds, keeping accepted implications, counterexamples, contradictions, and corrections inspectable.

This paper makes three contributions.
\begin{itemize}
  \item An FCA-based verification loop that tests object--attribute structures through implications and counterexamples.
  \item A retrieval-grounded SLM oracle for lower cost evidence based local judgments.
  \item A symbolic--subsymbolic hybrid where FCA controls construction and the language model makes evidence-based local decisions.
\end{itemize}
Together, the experiments position the method as a verifiable construction procedure rather than an unchecked ontology generator.
The ablations further separate the difficulty of discovering object and attribute sets from the difficulty of judging incidences within those sets.

\section{Background}
FCA supplies the symbolic side by representing the current construction state as a formal context and testing implications with counterexamples.
RAG supplies the evidence grounding side by retrieving disease level text for the SLM oracle, turning ontology construction into a verifiable loop of implication queries, local judgments, counterexamples, and context updates.

\subsection{Formal concept analysis and attribute exploration}

Formal concept analysis (FCA) represents object--attribute knowledge as a binary table called a formal context~\cite{ganter1999fca}.
In this paper, the objects are rare ataxia diseases and the attributes are phenotype labels.
A cell is marked when a disease is associated with a phenotype.
For example, one disease may have \textit{Ataxia}, \textit{Tremor}, and \textit{Dysarthria}, while another disease may have \textit{Ataxia} and \textit{Tremor} but not \textit{Dysarthria}.
A formal concept consists of an extent, the set of objects in the concept, and an intent, the attributes shared by those objects.
An implication describes a regularity in the table.
The implication $A \rightarrow B$ means that every object having all attributes in $A$ also has all attributes in $B$.
For example, $\{\textit{Ataxia},\textit{Tremor}\}\rightarrow\{\textit{Dysarthria}\}$ states that every disease with \textit{Ataxia} and \textit{Tremor} also has \textit{Dysarthria}.
Such a rule is useful only if it survives a counterexample check.
A disease with \textit{Ataxia} and \textit{Tremor} but without \textit{Dysarthria} would reject it.

Attribute exploration turns these regularities into oracle questions.
Following Algorithm~19 in \textit{Conceptual Exploration}~\cite{ganter2016conceptual}, the intended domain context $(G,M,I)$ is distinguished from the observed partial context $(E_\tau,M_\tau,J_\tau)$ at round $\tau$.
Here, $G$ is the domain object set, $M$ is the finite attribute set, $I$ is the target incidence relation, $E_\tau$ is the observed object set, $M_\tau$ is the active attribute set, and $J_\tau$ records checked incidences.
The closure $A^{J_\tau J_\tau}$ is the set of attributes shared by the currently observed objects that have all attributes in $A$.
The exploration query asks whether the observed implication $A\rightarrow A^{J_\tau J_\tau}$ also holds in the intended domain.
If the oracle accepts the query, the implication is added to the implication base.
If the oracle rejects it, the oracle must return a counterexample object $g$ such that $A\subseteq g^I$ but $A^{J_\tau J_\tau}\nsubseteq g^I$.
The counterexample is added to the observed context, preventing the same overgeneral rule from being accepted again.
Classical attribute exploration assumes that $M$ is fixed from the beginning; under a reliable expert and finite fixed $M$, it can return a canonical implication basis.
Following the FCA usage in \textit{Conceptual Exploration}, this term refers to a complete and non redundant implication basis of a fixed formal context.
Every valid implication of the context is derivable from the basis, and no accepted implication is treated as a free text rule outside the context.
In this paper, FCA and attribute exploration provide the symbolic mechanism for proposing implications, checking counterexamples, and updating the disease--phenotype context.

\subsection{Retrieval-augmented generation}

Retrieval-augmented generation addresses a central limitation of purely parametric language models.
Factual decisions should be grounded in external evidence that can be inspected or updated.
Lewis et al.~\cite{lewis2020rag} introduced RAG as a way to combine parametric generation with non parametric retrieval for knowledge intensive NLP tasks.
In ontology construction, retrieval is useful because object--attribute relations and implications require evidence sensitive judgments.
Without retrieved evidence, a language model must answer from parametric memory alone, which can produce unsupported or incorrect relations.
Retrieved evidence gives the model task relevant information to consult before making each local judgment.
Providing retrieved evidence can improve performance over relying on parametric generation alone, especially in knowledge grounded dialogue and other knowledge intensive tasks~\cite{shuster2021retrieval,huang2025hallucination}.
In this paper, retrieved evidence is used to ground local object--attribute and implication validity questions.

\subsection{Small language models}

Large language models remain advantageous for open ended generation and multi step reasoning.
However, the decisions required in the FCA loop are narrower than open ended generation.
They are repeated, evidence conditioned YES/NO judgments over local object--attribute assignments or candidate implications.
For this setting, maximum accuracy is not the only consideration.
Inference cost, latency, memory footprint, and local deployability also matter.
This task dependent view of model size is supported by prior work on text classification, where specialized smaller models can reach or exceed the performance of general large models with a limited number of labeled examples~\cite{pecher2025specialised}.
It is also supported by industrial studies showing that smaller transformer models can handle practical classification workloads while offering better deployment efficiency~\cite{li2025small}.
These findings match the use of SLMs for constrained classification style judgments over retrieved evidence rather than for long form reasoning or free form ontology generation.
Thus, SLMs are not a replacement for LLMs in complex reasoning, but they are a cost efficient option when the task can be reduced to repeated, evidence conditioned YES/NO decisions.

\section{Method}

This section describes how the symbolic and retrieval-grounded components are combined into a single ontology construction loop.
The framework starts from seed attributes and uses FCA attribute exploration to generate implication queries.
For each query, the system first retrieves disease level evidence relevant to the premise and conclusion.
The SLM oracle then uses this retrieved evidence to accept the implication or search for a counterexample.
The formal context is updated with accepted implications or counterexample objects.
Unlike classical attribute exploration, this setting keeps a finite controlled phenotype attribute set but activates it progressively.
The loop starts from seed attributes $M_0$, explores only the active attribute set $M_\tau$, and then adds a selected set of new attributes $\Delta M_\tau$ for the next round.
Each round keeps the FCA-based verification step.
The overall process tests whether a small seed set can expand into useful parts of $M$, while avoiding unchecked free text ontology generation.
The following paragraphs define the round state, context updates, oracle decisions, and logged artifacts.

\begin{figure*}
  \centering
  \includegraphics[width=\textwidth]{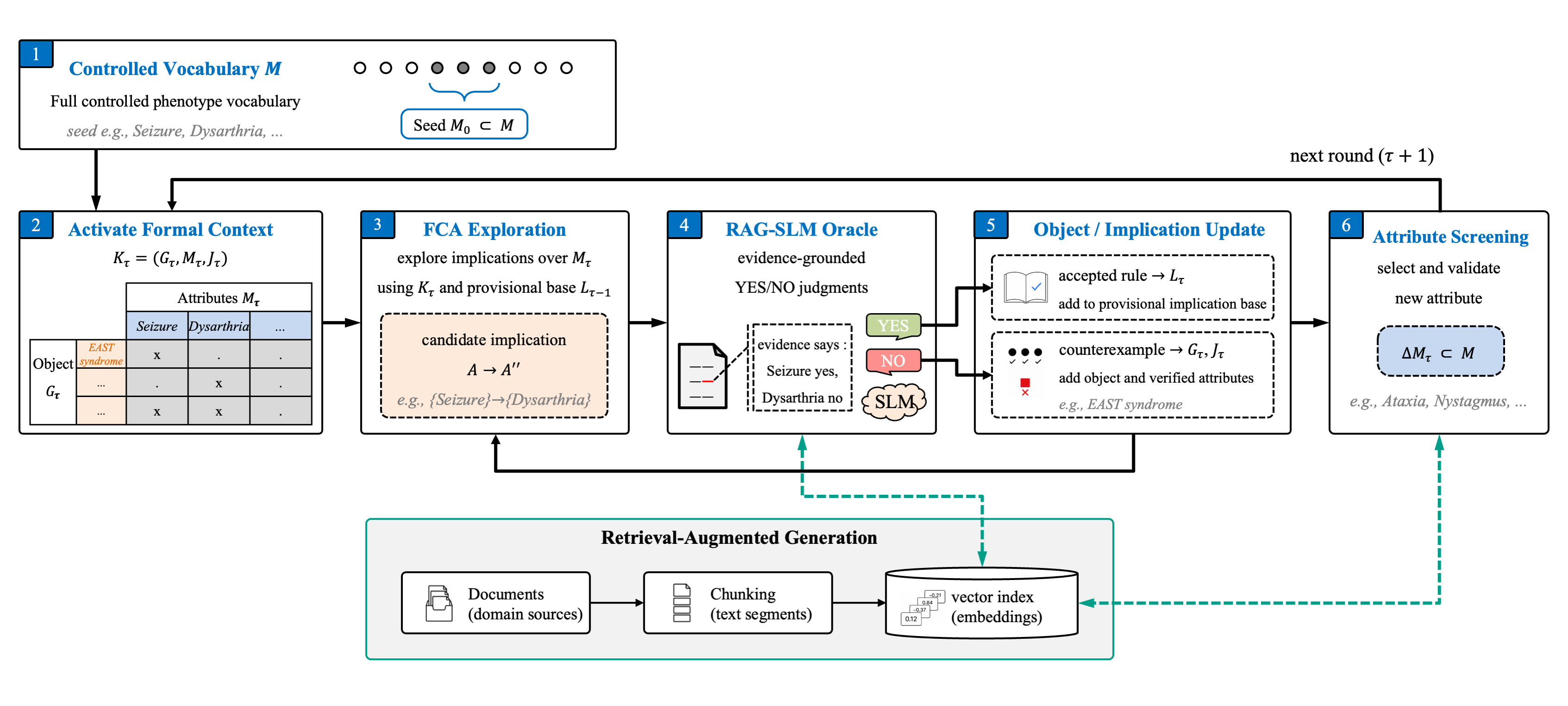}
  \caption{Overview of the RAG-grounded SLM-FCA framework.}
  \Description{A flow diagram of the RAG-grounded SLM-FCA framework, showing the FCA exploration loop, retrieval-grounded oracle decisions, counterexample validation, and implication-guided attribute discovery updates.}
  \label{fig:framework}
\end{figure*}

\subsection{RAG-grounded SLM-FCA framework}

The proposed method combines the FCA and RAG components introduced above into a single verifiable construction loop.
Figure~\ref{fig:framework} summarizes the loop as six stages.
The paragraphs below define the main components of this loop, from controlled attribute selection to attribute screening for the next round.

\textbf{Controlled attribute set.}
The method starts from a finite domain attribute set $M$ supplied before exploration and activates only a seed subset $M_0$ in the first round.
In Figure~\ref{fig:framework}, the term \emph{controlled vocabulary} denotes the finite domain attribute set \(M\), i.e., the full controlled set of phenotype attributes considered by the method.
The seed set is chosen to cover diverse, interpretable regions of the attribute set rather than a single narrow attribute cluster.
Good seeds should be evidence accessible, non redundant, and frequent enough to produce early counterexamples, while still specific enough to avoid accepting overly broad implications.
The rare ataxia experiment in Section~4 uses this criterion to choose clinically interpretable seed attributes.

\textbf{Activate formal context.}
The initial context contains seed attributes but no seed objects, and rejected initial queries introduce the first counterexamples.
At round $\tau$, the active state is the formal context $(E_\tau,M_\tau,J_\tau)$, where $E_\tau$ is the observed object set, $M_\tau$ is the active attribute set, and $J_\tau$ records checked incidences.

\textbf{FCA exploration.}
At round $\tau$, FCA attribute exploration runs over the current attribute set $M_\tau$ and example context $(E_\tau,M_\tau,J_\tau)$, producing candidate implications $A \rightarrow A^{J_\tau J_\tau}$.
Here, $J_\tau$ is the currently observed incidence relation, and $A^{J_\tau J_\tau}$ is the current closure of $A$, i.e., the attributes shared by all observed objects that have every attribute in $A$.
For readability, Algorithm~\ref{alg:rag_slm_fca} writes the same closure as $\operatorname{cl}_{K_\tau}(A)$, where $K_\tau=(E_\tau,M_\tau,J_\tau)$.
When an implication is not already closed in the current context, the retrieval-grounded oracle either returns a verified counterexample object or accepts the implication.
After attribute exploration finishes for the fixed $M_\tau$, the method selects a small set of new attributes for the next round.
Because $M_\tau$ grows from a seed, the output is an explored implication set over the activated context, not a claim of a complete canonical basis for the full controlled phenotype attribute set.

\textbf{RAG-SLM oracle.}
The oracle uses short prompts for binary YES/NO decisions over retrieved evidence.
These decisions cover object--attribute judgments, counterexample checks, and corrections.
Retrieval queries are built from the labels in the current FCA question.
For an object--attribute query, the query combines the disease name with the phenotype label being checked.
For an implication query, the query combines premise and conclusion labels, such as \textit{Ataxia}, \textit{Cerebellar atrophy}, and \textit{Dysarthria}.
The retrieved snippets are then placed in the oracle prompt as evidence for accepting the implication or searching for a counterexample.
When the loop needs additional attributes, the SLM proposes reusable property phrases from retrieved snippets, which are then filtered, validated, and checked before entering the context.
This prompt design follows prior findings that retrieval-augmented outputs are sensitive to context placement, filtering, and distracting evidence~\cite{ram2023incontext,wang2023filco,liu2024lost}.
The prompt therefore shows only the most relevant snippets near the YES/NO question before constrained output generation.

\textbf{Context and implication update.}
The symbol $\Delta M_\tau$ denotes the selected attributes appended after round $\tau$, not the entire attribute set.
Accepted implications are reused in later rounds as provisional background knowledge for query avoidance and contradiction detection.
When a later counterexample candidate conflicts with a previously accepted implication, the implementation re checks the relevant object--attribute incidence and logs the correction event.
It does not perform full implication retraction or belief revision.
Thus, the method is an inspectable construction procedure, not a complete expert verified ontology editor.

\textbf{Attribute screening.}
Attribute proposals are guided by the current implication structure.
For an accepted implication $A \rightarrow \operatorname{cl}_{K_\tau}(A)$, the closure difference $\operatorname{cl}_{K_\tau}(A)\setminus A$ supplies inferred attributes that can seed retrieval queries.
The system retrieves evidence from the premise, from the premise plus one inferred attribute, and from premise satisfying object anchors; if these queries fail, it visits unused vector database chunks one at a time.
Only short, grounded, non duplicate property phrases that can be judged as YES/NO attributes are mapped to canonical attribute labels, selected, and appended to $\Delta M_\tau$.

\textbf{Oracle decisions.}
The retrieval-grounded SLM oracle is used only for constrained local decisions.
For object--attribute queries, it returns a binary YES/NO incidence judgment from retrieved evidence, with unsupported or ambiguous evidence treated as NO.
For implication queries, it searches for a counterexample object that satisfies the premise but misses at least one conclusion attribute.
Accepted implications enter the implication base, while rejected implications add verified counterexample objects and checked incidences to the context.
Later consistency conflicts are recorded as logged contradictions and corrections rather than full implication retractions.

Algorithm~\ref{alg:rag_slm_fca} summarizes this round-level procedure in pseudocode.

\begin{algorithm}[H]
  \caption{RAG-grounded SLM-FCA ontology exploration}
  \label{alg:rag_slm_fca}
  \begin{algorithmic}[1]
    \Require Controlled attribute set $M$, seed attributes $M_0 \subseteq M$, retrieval index, SLM oracle
    \State Initialize formal context $K_0=(E_0,M_0,J_0)$ with $E_0=\varnothing$
    \For{round $\tau = 1,\dots,T$}
    \State Update object incidences for new attributes, then run FCA on $K_\tau$
    \For{each candidate implication $A \rightarrow \operatorname{cl}_{K_\tau}(A)$}
    \State Retrieve evidence relevant to $A$ and $\operatorname{cl}_{K_\tau}(A)$
    \State Ask the SLM oracle to search for and verify a counterexample
    \If{the implication is rejected}
    \State Add the counterexample object and checked incidences to $K_\tau$
    \Else
    \State Add $A \rightarrow \operatorname{cl}_{K_\tau}(A)$ to the implication base
    \EndIf
    \EndFor
    \State Use implications and object anchors to retrieve evidence and propose attributes
    \If{no attribute is selected from the pending pool}
    \State Visit unswept vector database chunks sequentially, one chunk at a time, and ask the SLM to extract candidate attributes
    \EndIf
    \State Validate, map to labels in $M$, and select $\Delta M_\tau$ for the next round
    \If{$\Delta M_\tau=\varnothing$}
    \State \textbf{break}
    \EndIf
    \State Set $M_{\tau+1}=M_\tau\cup\Delta M_\tau$ and carry objects forward
    \EndFor
    \State \Return final context and implication base
  \end{algorithmic}
\end{algorithm}

\subsection{Verification Artifacts}

Each round produces artifacts that make the construction process inspectable.
The system records the evolving object--attribute matrix, accepted implications, counterexamples, and logged contradiction/correction events from consistency checks.
It also records final evaluation outputs for disease--phenotype relation metrics and closure based implication metrics.
These artifacts are the basis for the paper's claim of verifiable ontology construction.
The output is therefore not only a final context, but also a trace of why structural commitments were accepted, rejected, or revised.

\section{Experiment}

This section evaluates whether the proposed framework can construct and verify a rare ataxia disease--phenotype context from Orphadata derived resources.
The setting separates disease level retrieval text used by the oracle from curated disease--HPO annotations used only for evaluation.

The experiments are run for 20 rounds with temperature 0.
The default 10-seed condition starts from frequent, clinically interpretable attributes covering speech, gait, cerebellar, eye-movement, pyramidal, and seizure-related findings.
The controlled exploration attribute set contains the 160 evidence-accessible HPO attributes described below, and the retrieval index embeds disease-aligned Orphanet records with EmbeddingGemma-300M\footnote{\url{https://huggingface.co/google/embeddinggemma-300m}}.
The same budgets are used across models, with up to ten counterexample candidates per implication test and up to three attribute discovery retrieval queries per round.
Gold disease--HPO relations are never used by the runtime oracle, and generated attributes are aligned to the gold attribute set only after exploration for evaluation.

Evaluation has two parts.
Predicted object--attribute relations are compared with the run specific projection of the evidence accessible gold context.
FCA implications induced by the output context are evaluated against the canonical basis of that context, so implication scores measure closure behavior rather than recovery of all implications in the full curated disease--phenotype annotation context.

\subsection{Dataset}

The framework is evaluated on rare ataxia diseases using Orphadata\footnote{\url{https://www.orphadata.com/}} and HPO~\cite{robinson2008hpo,kohler2021hpo}.
Orphadata provides disease identifiers, names, classification records, and disease definition text, while HPO provides phenotype labels and curated disease--phenotype relations.
In the resulting formal context, objects are rare ataxia diseases, attributes are HPO phenotype labels, and a positive cell means that the disease is curated as having that phenotype.

The disease objects are selected from the Rare ataxia branch of the Orphadata rare neurological disease classification.
From 189 classification nodes, 139 disease candidates are retained, 124 remain after requiring curated HPO annotations, and 122 remain after requiring aligned Orphanet disease definition text.
Thus, every object has curated phenotype annotations for evaluation and disease definition text for retrieval.

The phenotype attribute set is derived from the curated disease--HPO annotations of these 122 diseases, which contain 850 HPO labels and 2,716 positive relations.
Because the retrieval corpus contains disease definition text rather than the curated annotation table, the evidence accessible gold context is constructed by exact keyword matching.
Only HPO labels whose canonical names appear in the disease aligned retrieval chunks are retained, yielding 122 disease objects, 160 HPO phenotype attributes, and 1,398 positive relations.
This attribute level filtering means that some retained curated relations are not explicitly stated in the corresponding disease definition text and may not be recoverable from retrieval evidence alone.

The runtime oracle sees only disease definition text, and evaluation uses curated HPO disease--phenotype annotations.

\subsection{Main Results}

The evaluation reports relation scores over disease--phenotype incidences and closure-based implication scores over predicted implications.
Table~\ref{tab:orphanet-rag-profile} reports how far each 10-seed RAG-grounded run expands the formal context over 20 rounds.
The reported objects are discovered counterexamples rather than a fixed input disease list.
Thus, larger object counts indicate that the exploration loop found more diseases that challenged candidate implications.
In Tables~\ref{tab:orphanet-rag-profile} and~\ref{tab:orphanet-seed20-profile}, Obj., Attr., Impl., Eval., and Qry. denote objects, attributes, accepted implications, evaluated implications, and oracle queries.
Eval. counts accepted implications whose attributes all fall inside the evidence accessible gold context.
Accepted implications outside this context are not scored by the gold projection.

\begin{table}[H]
  \centering
  \caption{10-seed exploration profile.}
  \label{tab:orphanet-rag-profile}
  \small
  \setlength{\tabcolsep}{2pt}
  \begin{tabular*}{\columnwidth}{@{\extracolsep{\fill}}lrrrrr@{}}
    \toprule
    Model        & Obj. & Attr. & Impl. & Eval. & Qry. \\
    \midrule
    Gemma2:2B    & 18   & 46    & 99    & 22    & 115   \\
    Llama3.2:3B  & 44   & 68    & 298   & 43    & 444   \\
    Qwen2.5:1.5B & 36   & 36    & 263   & 11    & 285   \\
    \bottomrule
  \end{tabular*}
\end{table}

The models use these objects and newly selected attributes to move beyond the seed attribute set.
Table~\ref{tab:orphanet-rag-profile} therefore describes the exploration behavior, not only the final matrix size.
Llama expands most aggressively, reaching 44 objects, 68 attributes, 298 accepted implications, and 444 oracle queries.
Qwen adds almost as many counterexample objects as Llama but keeps a smaller attribute set, while Gemma remains the smallest explored object context.
The small Eval. counts show that many accepted implications include generated attributes that cannot be projected back to the gold attribute set.
Table~\ref{tab:orphanet-rag-metrics} evaluates the final contexts from the same runs against the evidence accessible gold context.
Rel. denotes relation scores, Impl. denotes closure based implication scores, and bold and underline mark the best and second best values within a table.

\begin{table}[H]
  \centering
  \caption{10-seed RAG-grounded SLM-FCA evaluation.}
  \label{tab:orphanet-rag-metrics}
  \small
  \setlength{\tabcolsep}{0pt}
  \begin{tabular*}{\columnwidth}{@{\extracolsep{\fill}}lrrrrrr@{}}
    \toprule
    Model        & Rel.P              & Rel.R              & Rel.F1             & Impl.P             & Impl.R             & Impl.F1            \\
    \midrule
    Gemma2:2B    & 0.18               & \textbf{0.87}      & 0.29               & \underline{0.23}   & 0.21               & 0.22               \\
    Llama3.2:3B  & \textbf{0.58}      & 0.47               & \textbf{0.52}      & 0.16               & \textbf{0.44}      & \underline{0.24}   \\
    Qwen2.5:1.5B & \underline{0.34}   & \underline{0.74}   & \underline{0.46}   & \textbf{0.27}      & \underline{0.34}   & \textbf{0.30}      \\
    \bottomrule
  \end{tabular*}
\end{table}

Table~\ref{tab:orphanet-rag-metrics} shows complementary metric tradeoffs.
Llama has the best relation F1 (0.52) and relation precision (0.58).
Qwen has the best implication F1 (0.30), while Gemma has the highest relation recall (0.87).
Overall, the runs construct verifiable partial contexts, while relation agreement and implication quality diverge sharply.

To examine the effect of a larger seed set, Tables~\ref{tab:orphanet-seed20-profile} and~\ref{tab:orphanet-seed20-metrics} report the corresponding 20-seed runs under the same 20-round budget.
The 10-seed runs in Tables~\ref{tab:orphanet-rag-profile} and~\ref{tab:orphanet-rag-metrics} provide the comparison point.
The 20-seed setting expands contexts and increases evaluable implications.

\begin{table}[H]
  \centering
  \caption{20-seed exploration profile.}
  \label{tab:orphanet-seed20-profile}
  \small
  \setlength{\tabcolsep}{2pt}
  \begin{tabular*}{\columnwidth}{@{\extracolsep{\fill}}lrrrrr@{}}
    \toprule
    Model        & Obj. & Attr. & Impl. & Eval. & Qry. \\
    \midrule
    Gemma2:2B    & 21   & 53    & 172   & 57    & 183   \\
    Llama3.2:3B  & 55   & 72    & 404   & 98    & 517   \\
    Qwen2.5:1.5B & 58   & 49    & 544   & 67    & 577   \\
    \bottomrule
  \end{tabular*}
\end{table}

The 20-seed setting increases accepted implications for all three models, from 99 to 172 for Gemma, from 298 to 404 for Llama, and from 263 to 544 for Qwen.
It also increases evaluable implications from 22 to 57 for Gemma, from 43 to 98 for Llama, and from 11 to 67 for Qwen.
Table~\ref{tab:orphanet-seed20-metrics} then evaluates whether this larger explored space improves relation and implication quality.

\begin{table}[H]
  \centering
  \caption{20-seed RAG-grounded SLM-FCA evaluation.}
  \label{tab:orphanet-seed20-metrics}
  \small
  \setlength{\tabcolsep}{0pt}
  \begin{tabular*}{\columnwidth}{@{\extracolsep{\fill}}lrrrrrr@{}}
    \toprule
    Model        & Rel.P              & Rel.R              & Rel.F1             & Impl.P             & Impl.R             & Impl.F1            \\
    \midrule
    Gemma2:2B    & 0.20               & \textbf{0.86}      & 0.32               & \underline{0.42}   & 0.31               & \underline{0.36}   \\
    Llama3.2:3B  & \textbf{0.56}      & 0.40               & \textbf{0.47}      & 0.15               & \textbf{0.73}      & 0.25               \\
    Qwen2.5:1.5B & \underline{0.32}   & \underline{0.64}   & \underline{0.42}   & \textbf{0.57}      & \underline{0.32}   & \textbf{0.41}      \\
    \bottomrule
  \end{tabular*}
\end{table}

The larger seed set improves implication F1 for all three models.
The largest absolute gain is Gemma, rising from 0.22 to 0.36, while Qwen reaches the highest final implication F1 at 0.41.
The gap between accepted and evaluable implications in Table~\ref{tab:orphanet-seed20-profile} remains large.
Many generated attributes still fall outside the matched gold attribute set.
Thus, the larger seed set improves reach but does not remove the evaluability bottleneck.

\begin{figure*}[!b]
  \centering
  \includegraphics[width=.94\textwidth]{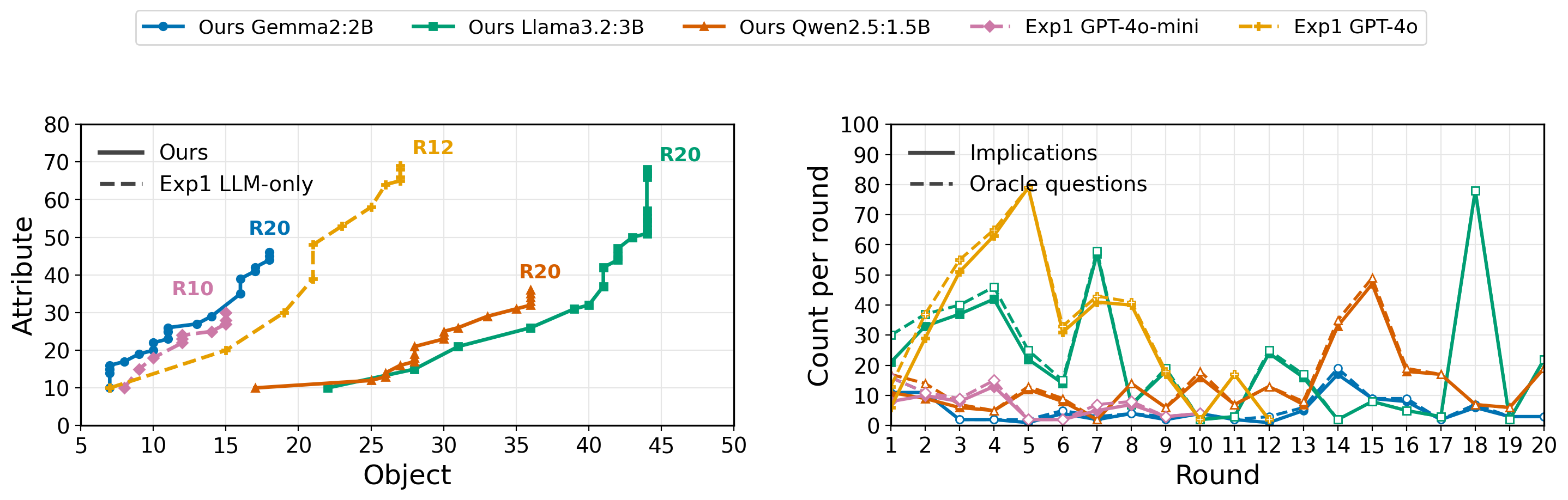}
  \caption{Round-level context growth and oracle activity. The left panel shows object--attribute expansion; the right panel shows accepted implications and oracle questions per round.}
  \Description{Two line charts comparing RAG-grounded SLM-FCA runs and LLM-only FCA Experiment 1 runs. The left chart shows object and attribute growth across rounds. The right chart shows accepted implications and oracle questions per round.}
  \label{fig:round-dynamics}
\end{figure*}

For closure based implication evaluation, a canonical implication basis is first computed from the evidence accessible gold context.
This basis is stored as the reference implication basis.
In the sense of classical FCA, this basis is a complete implication representation of that fixed formal context.
Valid implications of the context are derivable from the basis by closure.
The same closure based implication scoring protocol is then applied to every run.
For implication precision, a predicted implication $A \rightarrow B$ is counted as correct when $B$ is contained in the closure of $A$.
This closure is computed under the canonical basis of the evidence accessible gold context.
Equivalently, the implication is entailed if every gold disease with the premise phenotypes in $A$ also has the predicted conclusion phenotypes in $B$.
For example, $A \rightarrow \{\text{Dysarthria}\}$ is correct when Dysarthria belongs to the gold closure of $A$, even if that closure contains additional phenotypes.
Thus, a prediction may state only part of the gold closure and still be correct; it is not required that the predicted implication exactly match a stored gold implication string.
For implication recall, each gold basis implication $C \rightarrow D$ is counted as recovered when $D$ is contained in the closure of $C$ under the predicted implication set.
This closure based implication protocol is used for the proposed framework and for Experiments~1--3 in the ablation study.
Table~\ref{tab:orphanet-implication-examples} gives concrete implication examples from the final run outputs.
Both examples are taken from the final 20-seed RAG-grounded run outputs.
The table keeps only the judgment type and implication so the examples remain readable.

\begin{table}[H]
  \centering
  \caption{Example implication judgments.}
  \label{tab:orphanet-implication-examples}
  \begingroup
  \small
  \setlength{\tabcolsep}{3pt}
  \begin{tabular}{l>{\raggedright\arraybackslash}p{0.60\columnwidth}}
    \toprule
    Case              & Example                                                                 \\
    \midrule
    Entailed (TP)     & \{Ataxia, Progressive cerebellar ataxia\} $\rightarrow$ \{Gait ataxia\} \\
    Non-entailed (FP) & \{Ataxia, Gait disturbance\} $\rightarrow$ \{Dysarthria, Gait ataxia\}  \\
    \bottomrule
  \end{tabular}
  \endgroup
\end{table}

The core FCA output in this seeded setting is an explored implication set.
The table contrasts predicted implications that are entailed by the evidence accessible gold context with those that are not.
In the Case column, \textit{Entailed (TP)} means that the predicted implication is true under the gold closure.
\textit{Non-entailed (FP)} means that the run predicted the implication, but the gold closure does not license it.
These non entailed examples are not necessarily clinically implausible.
They are simply not supported by the evidence accessible gold context under closure based evaluation.

The same logs also show how the system makes verification failures inspectable.
During exploration, a failed implication introduces a disease level counterexample rather than silently accepting a plausible rule.
During consistency checking, an existing object can violate an accepted implication.
The system records the violated premise, missing conclusion attributes, and any corrected disease--phenotype relations proposed from the same Orphanet nomenclature evidence bundle.

\paragraph{Trace examples.}
In the 20-seed runs, Gemma2:2B records 21 disease level counterexamples, 57 logged contradictions, and 6 corrections.
Llama3.2:3B records 55 counterexamples, 132 logged contradictions, and no accepted corrections.
The corresponding Qwen2.5:1.5B run records 58 counterexamples, 105 logged contradictions, and 8 corrections.
These traces show that failed implications are not hidden.
They become explicit counterexample objects, violation records, and candidate corrections that can be inspected after the run.
The contrast also explains the metric pattern in Table~\ref{tab:orphanet-seed20-metrics}.
Llama expands the context most aggressively, while Qwen produces stronger implication F1 with moderate relation recall and lower precision.

\subsection{Ablation Study}
Tables~\ref{tab:orphanet-exp1-results}--\ref{tab:orphanet-exp3-results} separate three GPT-based ablations from the local SLM oracle used in the proposed framework.

\textbf{Experiment~1.} Retrieval is removed from the exploration loop.
GPT-4o-mini or GPT-4o is used as the oracle.
Counterexample validation, consistency checking, and attribute expansion are retained, with GPT providing attribute proposals.
The runs stop when the exploration loop reaches its termination condition rather than a fixed round limit.
Table~\ref{tab:orphanet-exp1-results} tests whether a GPT-based oracle can compensate for removing retrieved disease evidence.

\begin{table}[H]
  \centering
  \caption{LLM-only FCA loop.}
  \label{tab:orphanet-exp1-results}
  \small
  \setlength{\tabcolsep}{0pt}
  \begin{tabular*}{\columnwidth}{@{\extracolsep{\fill}}lrrrrrrr@{}}
    \toprule
    Oracle & Rnd & Obj. & Attr. & Impl. & Qry. & Rel.F1             & Impl.F1            \\
    \midrule
    GPT-4o-mini  & 10  & 15   & 30    & 62    & 77   & \textbf{0.56}      & \textbf{0.67}      \\
    GPT-4o       & 12  & 27   & 69    & 378   & 405  & \underline{0.54}   & \underline{0.41}   \\
    \bottomrule
  \end{tabular*}
\end{table}

These results show that GPT-based LLM-only runs can achieve high implication quality in smaller explored contexts, but they do not replace retrieval driven context growth.
Because the LLM-only runs terminate early, their high implication F1 is diagnostic rather than evidence that retrieval is unnecessary or directly comparable to larger retrieved contexts.
This separates oracle strength from the evidence supply that helps the loop find counterexamples and attributes.

Figure~\ref{fig:round-dynamics} adds a round level view of the difference between retrieval-grounded exploration and the LLM-only ablation.
The left panel shows that the RAG-grounded runs continue to add objects and attributes across rounds, whereas the LLM-only runs saturate after fewer rounds.
The right panel shows that oracle questions and accepted implications are not the same signal.
Many oracle calls are used for verification, incidence checks, or counterexample search rather than becoming accepted implications.
Together, these trends show that retrieved disease evidence sustains exploration by supplying new object and attribute evidence, while a GPT-based oracle without retrieval tends to operate within a narrower explored context.

\textbf{Experiment~2.} This experiment uses the disease objects and phenotype attributes from each 10-seed, 20-round RAG-grounded source run in Table~\ref{tab:orphanet-rag-profile}.
For Gemma2, Llama3.2, and Qwen2.5, these discovered objects and attributes are reused as the candidate matrix.
No new diseases or phenotype labels are added.
GPT-4o-mini then rejudges every disease--phenotype cell in each source model's matrix.
The completed matrix is evaluated with the same projected implication protocol.
Thus, Table~\ref{tab:orphanet-exp2-results} isolates incidence quality after object and attribute discovery has already happened.

\begin{table}[H]
  \centering
  \caption{Incidence judgment on discovered object--attribute matrices.}
  \label{tab:orphanet-exp2-results}
  \small
  \setlength{\tabcolsep}{0pt}
  \begin{tabular*}{\columnwidth}{@{\extracolsep{\fill}}lrrrrr@{}}
    \toprule
    Source & Obj. & Attr. & Qry. & Rel.F1             & Impl.F1            \\
    \midrule
    Gemma2:2B       & 18   & 46    & 828   & 0.39               & \underline{0.28}   \\
    Llama3.2:3B     & 44   & 68    & 2,992 & \textbf{0.47}      & 0.26               \\
    Qwen2.5:1.5B    & 36   & 36    & 1,296 & \underline{0.46}   & \textbf{0.40}      \\
    \bottomrule
  \end{tabular*}
\end{table}

The rejudged matrices indicate that incidence judgment remains a bottleneck even after the source model has found the candidate objects and attributes.
Relation F1 stays below 0.50 for all three matrices, with Llama highest and Qwen close behind.
The induced implication scores improve over the original SLM oracle judgments for all three 10-seed matrices, with the largest gain for Qwen.
Thus, local incidence decisions affect structural quality, but rejudged cell decisions do not uniformly improve every relation metric.

\textbf{Experiment~3.} The candidate diseases and phenotype labels are taken from the full evidence accessible gold context.
This fixed object--attribute setting is kept unchanged throughout the experiment.
GPT-4o-mini then judges every disease--phenotype pair within that fixed evaluation setting.
The completed incidence table is evaluated for both relation quality and induced implications.
Table~\ref{tab:orphanet-exp3-results} reports this favorable fixed object--attribute setting over 122 diseases and 160 phenotype labels, yielding $19{,}520$ disease--phenotype pairs.
\begin{table}[H]
  \centering
  \caption{Incidence judgement on the fixed gold object--attribute matrices.}
  \label{tab:orphanet-exp3-results}
  \small
  \setlength{\tabcolsep}{0pt}
  \begin{tabular*}{\columnwidth}{@{\extracolsep{\fill}}lrrrrrr@{}}
    \toprule
    Model       & Rel.P & Rel.R & Rel.F1 & Impl.P & Impl.R & Impl.F1 \\
    \midrule
    GPT-4o-mini & 0.22  & 0.54  & 0.31   & 0.91   & 0.36   & 0.51     \\
    \bottomrule
  \end{tabular*}
\end{table}

The fixed object--attribute setting shows that even the complete evaluation scope does not make incidence recovery easy.
The relation F1 remains only 0.31, so many disease--phenotype cells are still hard to judge from the available textual evidence.
This shows that the limitation is not only missing objects or missing phenotype labels.
The evidence text itself can be too indirect for reliable cell level recovery.
Because the discovery problem is removed, the model only has to judge whether each known pair is supported.
At the implication level, the resulting context achieves high precision but lower recall.
This precision--recall gap suggests conservative output: fewer gold implications are recovered, but many accepted ones remain valid.
Thus, the result is a favorable fixed object--attribute baseline rather than evidence that open world exploration has been solved.
The remaining errors show that relation recovery is still difficult even when all candidate diseases and phenotype attributes are already fixed.

\textbf{Cost.} Table~\ref{tab:orphanet-openai-cost} reports API calls and cost for the GPT-4o and GPT-4o-mini ablations.
The table separates model price from the number of oracle judgments.
Exp1 uses relatively few calls because the explored contexts are small, but its GPT-4o condition remains expensive because each call has a higher unit price.
Exp3 shows the opposite pattern because it uses GPT-4o-mini while evaluating every disease--phenotype pair in the evidence-accessible gold context.
Thus, local incidence judgments become the main driver of API usage.

\begin{table}[H]
  \centering
  \caption{OpenAI API usage.}
  \label{tab:orphanet-openai-cost}
  \small
  \setlength{\tabcolsep}{0pt}
  \begin{tabular*}{\columnwidth}{@{\extracolsep{\fill}}llrr@{}}
    \toprule
    Experiment    & Model       & Calls & Cost \\
    \midrule
    Exp1          & GPT-4o-mini & 77    & \$0.12 \\
    Exp1          & GPT-4o      & 405   & \$6.56 \\
    Exp2 Gemma2   & GPT-4o-mini & 828    & \$0.02 \\
    Exp2 Llama3.2 & GPT-4o-mini & 2,992  & \$0.06 \\
    Exp2 Qwen2.5  & GPT-4o-mini & 1,296  & \$0.03 \\
    Exp3          & GPT-4o-mini & 19,520 & \$0.38 \\
    \bottomrule
  \end{tabular*}
\end{table}

The cost table shows that API usage depends on both oracle strength and evaluation scope.
Exp2 calls follow the size of each discovered object--attribute matrix, while Exp3 grows to 19,520 calls because every fixed gold context pair is judged.

\subsection{Evaluation and Limitations}

The results characterize the framework as verifiable partial construction, not full ontology reconstruction.
Its output is an explored implication set rather than a canonical basis of the full controlled phenotype attribute set.
For the completed 10-seed RAG-grounded runs, relation F1 ranges from 0.29 to 0.52 and closure based implication F1 ranges from 0.22 to 0.30.
Experiment~1 reaches 0.56 relation F1 and 0.67 implication F1 in a smaller explored context.
These results demonstrate a working verification loop, but not high-precision ontology completion.
Part of this gap comes from the larger explored object--attribute contexts, where more discovered objects, selected attributes, and incidence decisions create more opportunities for cell level errors.
Because the retrieval corpus uses disease definition text rather than the curated annotation table, many HPO associations are not explicitly verbalized in the raw evidence.
This mismatch especially affects relation recall, since a curated phenotype can be valid even when its exact label is absent from the retrieved disease description.
It also affects implication scores indirectly, because a small number of missed or extra incidences can change several closure judgments.
The appropriate claim is therefore an inspectable construction process whose scores measure recovered incidences and closure behavior in the evidence accessible gold context.

\section{Discussion}

The results clarify why ontology construction should be treated as verified structure building, not unconstrained text generation.
FCA is the structural control mechanism because it derives implications from the current object--attribute context and tests them through counterexamples.
Retrieval supplies disease level evidence, and the SLM oracle provides a lower cost interface for repeated evidence conditioned decisions.
The GPT-based LLM-only ablation reaches high evaluation scores in smaller explored contexts, but Figure~\ref{fig:round-dynamics} shows that retrieval-grounded runs keep expanding objects and attributes.
Thus, final evaluation scores and evidence-driven context growth capture different aspects of the construction process.

The main limitation is not only the model.
The available disease definition text can also be too sparse for cell level incidence recovery.
Experiment~3 fixes the candidate diseases and phenotype attributes, but relation F1 remains low.
This suggests that some curated HPO relations cannot be recovered from disease definition text alone.
For this reason, the output should be read as an auditable pre curation artifact.
The trace records accepted implications, rejected implications, counterexample objects, contradiction logs, and corrections.
These records show where the construction succeeded or failed.
An expert can inspect them before accepting the resulting object--attribute context as ontology content.
This keeps the claim centered on inspectable construction rather than automatic ontology completion.

\section{Conclusion}

This paper presented a RAG-grounded SLM-FCA framework for verifiable ontology construction in rare ataxia phenotyping.
The framework starts from seed attributes and expands the context step by step.
FCA proposes structural commitments, retrieval provides disease evidence, and SLMs make repeated local judgments over that evidence.
Experiments on Orphadata-derived disease records show that the loop can add objects and attributes while recording counterexamples and corrections.
The logs also connect implications to object level evidence, which makes the construction process inspectable.
The ablations show that high scores in a fixed or smaller explored context do not describe the whole open world task.
In open world construction, the system must also discover useful objects and attributes.
The output should therefore be read as an explored implication set over an evidence accessible context.
It is not a finished clinical ontology or a canonical basis for the full controlled phenotype attribute set.
The central takeaway is that FCA can make retrieval-grounded language model construction inspectable.
Reliable ontology completion will still require stronger retrieval coverage, better synonym and entity matching for HPO labels, improved seed selection, stricter implication acceptance criteria, and expert facing review of logged oracle questions, counterexamples, and corrections.

\bibliographystyle{ACM-Reference-Format}
\bibliography{epiDAMIK}

\end{document}